\documentclass[10pt,letterpaper,twocolumn]{article}
\usepackage[margin=0.72in,columnsep=0.25in]{geometry}
\usepackage{fontspec}
\usepackage[english]{babel}
\usepackage{amsmath,amssymb,mathtools,unicode-math}
\usepackage{algorithm}
\usepackage{algpseudocode}
\usepackage{graphicx,booktabs,tabularx,array,seqsplit}
\usepackage{multirow}
\usepackage{enumitem,etoolbox,needspace,float,placeins,xcolor}
\usepackage[font=small,labelfont=bf,labelsep=period]{caption}
\usepackage[numbers,sort&compress]{natbib}
\usepackage[unicode,hidelinks,breaklinks]{hyperref}
\usepackage{bookmark,xurl,fancyhdr}
\usepackage[nospread,noshrink]{flushend}
\defaultfontfeatures{Ligatures=TeX}

\definecolor{draftorange}{HTML}{9C4D10}

\setlist{topsep=2pt,itemsep=1pt,parsep=0pt,leftmargin=*}

\AtBeginEnvironment{thebibliography}{\fontsize{8.5}{10.2}\selectfont\setlength{\bibsep}{2pt}\interlinepenalty=10000}

\makeatletter
\renewcommand\section{\@startsection{section}{1}{\z@}{-2.3ex \@plus -.8ex \@minus -.2ex}{.8ex \@plus .2ex}{\normalfont\large\bfseries}}
\renewcommand\subsection{\@startsection{subsection}{2}{\z@}{-1.6ex \@plus -.6ex \@minus -.2ex}{.5ex \@plus .2ex}{\normalfont\normalsize\bfseries}}
\renewcommand{\maketitle}{\twocolumn[{\begin{minipage}{\textwidth}\centering
{\fontsize{17}{20}\selectfont\bfseries\@title\par}\vspace{7pt}
{\fontsize{10.5}{12.5}\selectfont\@author\par}\vspace{3pt}
{\small Nanjing University\par}\vspace{2pt}
{\footnotesize\textsuperscript{†}Co-first authors.\quad
 \textsuperscript{*}Corresponding author.\par}\vspace{3pt}
{\small\url{https://nju3dv-humanoidgroup.github.io/CoAgent3D/}\par}\vspace{6pt}
\normalfont\normalsize\captionsetup{hypcap=false}
\includegraphics[width=\textwidth]{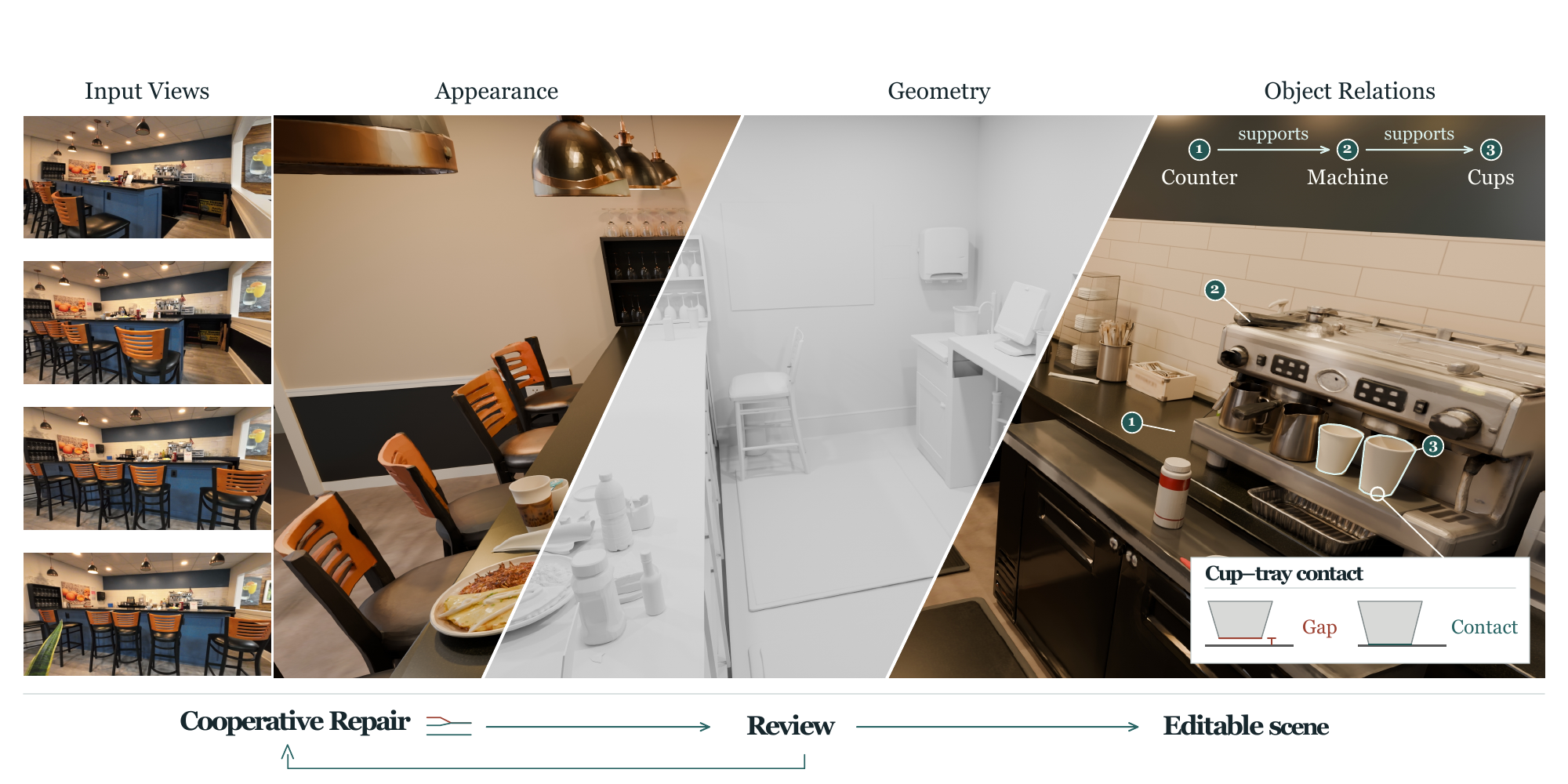}
\captionof{figure}{\textbf{CoAgent3D reconstructs photorealistic, editable 3D scenes from input images through relation-guided cooperative repair.}
Agents prepare repairs in parallel with full-scene context and coordinate coupled edits through shared decisions supported by an evolving relation graph.
Render comparisons and targeted mesh cross-sections guide iterative repair and independent review, refining appearance and object geometry while resolving support and contact inconsistencies.}\label{fig:teaser}
\vspace{8pt}
\end{minipage}}]\thispagestyle{plain}}
\makeatother
\algrenewcommand\textproc{\textnormal}

\title{CoAgent3D: Relation-Guided Cooperative Repair for Agentic 3D Scene Reconstruction}
\author{Ziyang Huang\textsuperscript{†}\quad
        Kaiming Xu\textsuperscript{†}\quad
        Yizhao Han\quad Xuan Yang\quad
        Xiaoxiao Long\textsuperscript{*}}
\hypersetup{pdfauthor={Ziyang Huang, Kaiming Xu, Yizhao Han, Xuan Yang, Xiaoxiao Long},pdftitle={CoAgent3D: Relation-Guided Cooperative Repair for Agentic 3D Scene Reconstruction}}
\begin{document}
\maketitle
\begin{abstract}
\small
Reconstructing complete, editable 3D scenes from visual observations is a fundamental problem in spatial intelligence.
Conventional reconstruction methods recover geometry and appearance from available views, leaving occluded structures underconstrained.
Recent agentic approaches introduce generative priors and tool use, but achieving photorealistic fidelity while coordinating edits across interdependent objects remains challenging.
We present \textbf{CoAgent3D}, a relation-guided self-improvement framework that organizes parallel scene repair around dependencies between proposed edits.
CoAgent3D builds an object-level relation graph from input images, assembles an initial scene, and identifies discrepancies through comparison with the observations.
Every repair agent retains access to the complete scene and publishes its repair intent, including proposed changes, affected elements, and unresolved assumptions.
The relation graph connects these intents to spatial requirements and supporting evidence.
Agents whose repairs depend on shared decisions exchange evidence and develop joint repairs, while independent preparation proceeds in parallel.
The resulting repairs are composed and undergo local and whole-scene review using mesh measurements, targeted cross-sections, and rendered comparisons.
Review outcomes update scene relations and unresolved issues, driving successive rounds of diagnosis and repair.
Across real-world scenes from multiple datasets, CoAgent3D produces photorealistic reconstructions with complete, editable object geometry, textures, and scene lighting.\par
\end{abstract}
% Authored section: introduction
\section{Introduction}\label{sec:introduction}

Reconstructing the 3D world from visual observations is a central problem in spatial intelligence.
Established approaches, including multi-view stereo (MVS)~\citep{schonberger2016pixelwise}, neural radiance fields (NeRF)~\citep{mildenhall2020nerf}, and 3D Gaussian splatting (3DGS)~\citep{kerbl2023gaussian}, recover scene geometry and appearance through multi-view geometry and image formation.
Their reconstructions depend on observation coverage, leaving occluded and unobserved surfaces underconstrained.
Recovering a complete, editable scene therefore requires reasoning about the objects and spatial relations underlying the images.
We study this problem for a specific observed scene: reconstructing its object identities, spatial arrangement, materials, and lighting to support object-level editing and photorealistic rendering.

Recent advances in multimodal models have enabled agents to combine visual reasoning, generative models, and tool use to construct and refine 3D scenes~\cite{xu2026aha3d,huang2026literealityagent}.
In complex real scenes, this refinement involves edits whose effects extend across object boundaries.
Moving a chair away from a table leg changes its clearance from neighboring chairs; modifying a support surface affects the objects resting on it.
These interactions make scene repair both a modeling problem and a coordination problem.
Agents responsible for different issues often need to agree on a common geometric interpretation before completing their own edits.

Three common designs expose a tension between parallelism, scene context, and edit compatibility.
A single agent retains complete context and coordinates its own changes, but serial repair makes complex scenes costly to refine.
Multiple agents with access to the complete scene can prepare repairs concurrently, yet produce conflicting changes to shared objects, boundaries, or spatial assumptions.
Partitioning the scene reduces these interactions, but also removes context needed to reason about neighboring supports, occluders, and geometry.
Effective collaboration must preserve complete scene context while coordinating the decisions that make repairs interdependent.

Existing systems organize collaboration through several mechanisms.
GTA propagates global requirements through recursively organized groups, while SceneMosaic provides cross-unit constraints for local Critic--Actor evolution~\cite{tao2026gta,ran2026scenemosaic}.
SceneSmith separates design, critique, and orchestration, and SceneAnything provides neighboring context to component agents and evaluates revisions in the assembled scene~\cite{pfaff2026scenesmith,xue2026sceneanything}.
Iterative repair introduces a further challenge: dependencies emerge as agents interpret observations and develop concrete edits.
Routing each newly discovered dependency through a central coordinator creates repeated replanning, while anticipating all interactions before repair begins is impractical for complex scenes.
Collaboration must therefore adapt as repair plans take shape.

We introduce \textbf{CoAgent3D}, a framework for relation-guided self-improvement in editable 3D scene reconstruction.
Our key idea is to organize cooperation around the dependencies of proposed repairs while giving every agent access to the complete scene.
Task assignments specify responsibility for resolving issues without imposing exclusive spatial editing boundaries.
Starting from an object-level relation graph and an initial scene assembly, agents identify discrepancies against the input observations and prepare repairs in private scene workspaces.
During preparation, they publish \textbf{repair intents} that describe proposed changes, affected scene elements, supporting evidence, and questions requiring input from other agents.
The relation graph provides a common spatial reference for this exchange, connecting proposed edits to the requirements supported by the observations.

Agents affected by a shared requirement negotiate directly to establish \textbf{shared decisions} and develop \textbf{joint repairs} for coupled issues.
For example, repairing a wooden frame may require space currently occupied by an adjacent cabinet.
The agents responsible for the frame and cabinet inspect the same observations, agree on the common boundary, and prepare the corresponding geometry changes together.
Other agents continue preparing repairs that do not depend on this decision.
Shared decisions remain linked to the repairs that adopt them, so changes to a common premise trigger re-examination and revision of the affected plans.
This mechanism coordinates coupled edits during preparation and preserves parallel progress elsewhere in the scene.

We then compose the repair programs into a candidate scene and conduct local and whole-scene review.
Mesh measurements, targeted cross-sections, and comparisons between rendered and observed images reveal errors in orientation, support, contact, and interpenetration.
Review outcomes update the relation graph, retain unresolved issues, and identify targets for the next round.
This cycle of diagnosis, collaborative repair, and review progressively refines the scene and the evidence guiding its reconstruction.
Across real-world scene datasets, CoAgent3D produces photorealistic reconstructions with editable object geometry, textures, and lighting.

Our contributions are threefold:
\begin{enumerate}
    \item \textbf{Parallel repair with complete scene context.}
    We organize concurrent repair preparation around scene relations and proposed edits while allowing every agent to reason over the complete scene.
    This preserves the spatial context required to address issues whose effects cross object boundaries.

    \item \textbf{Direct negotiation and joint repair.}
    We enable agents responsible for interdependent issues to exchange evidence, establish shared geometric decisions, and jointly prepare repairs.
    Explicit dependencies between decisions and repair plans propagate changes to the affected agents while independent preparation continues.

    \item \textbf{Relation-guided diagnosis, repair, and review.}
    We integrate mesh measurements, targeted cross-sections, and visual comparisons into an iterative reconstruction process supported by an evolving relation graph.
    Local and whole-scene review update relations, evidence, and unresolved issues to guide subsequent repairs.
\end{enumerate}
\section{Related Work}\label{sec:related}

Editable scene reconstruction draws on compositional modeling, multi-agent collaboration, and iterative feedback.
We review these three directions, focusing on how scene relations and visual evidence support reconstruction and coordinate subsequent edits.

\subsection{Compositional reconstruction and scene relations}\label{compositional-reconstruction-and-physical-scene-recovery}\label{scene-graphs-and-persistent-feedback}

Compositional methods recover explicit objects together with their spatial arrangement.
Total3DUnderstanding jointly estimates room layout, object boxes, and object meshes from a single image~\citep{nie2020total3d}.
HoloScene combines object relations, visual observations, physical constraints, and generative priors to reconstruct scenes from video~\citep{xia2025holoscene}.
ReplicateAnyScene aligns semantic descriptions, visual evidence, and spatial estimates for compositional reconstruction~\citep{dong2026replicateanyscene}, while SimRecon uses view selection for object completion and a synthesized scene graph for simulator assembly~\citep{xia2026simrecon}.
Scene graphs provide a structured representation of these relationships: Armeni et al.\ connect objects, rooms, and cameras within a building~\citep{armeni2019scenegraph}, and SceneGraphFusion incrementally updates a semantic graph from RGB-D observations~\citep{wu2021scenegraphfusion}.

CoAgent3D maintains an evolving relation graph that links requirements inferred from the observations to measurements of the current reconstruction.
This graph provides a common spatial reference for diagnosing issues and coordinating edits across objects.

\subsection{Multi-agent scene generation and repair}\label{collaborative-scene-generation-and-revision}

Multi-agent systems coordinate scene construction through task decomposition and specialized roles.
GTA propagates global requirements through recursively defined groups and processes independent groups in parallel~\citep{tao2026gta}.
SceneSmith assigns design, evaluation, and orchestration to separate agents across construction stages~\citep{pfaff2026scenesmith}.
SceneMosaic evolves layouts within anchor-centered units, incorporating cross-unit constraints and validation after composition~\citep{ran2026scenemosaic}.
SceneAnything supplies component agents with neighboring geometry and support constraints, then evaluates revisions through programmatic checks and independent visual comparisons in the assembled scene~\citep{xue2026sceneanything}.

CoAgent3D organizes cooperation around dependencies that emerge during repair preparation.
Every repair agent retains access to the complete scene and publishes repair intents.
Agents negotiate shared decisions when their proposals interact and prepare joint repairs for coupled changes, while independent work continues in parallel.
The system records the decision versions adopted by each repair, identifying which plans require rechecking when a shared decision changes.

\subsection{Agentic reconstruction and iterative feedback}\label{agentic-reconstruction-with-visual-and-geometric-feedback}

Iterative feedback improves generated outputs through repeated evaluation and revision, as exemplified by Self-Refine~\citep{madaan2023selfrefine}.
In 3D authoring, BlenderAlchemy selects Blender program edits through rendering and visual evaluation~\citep{huang2024blenderalchemy}.
AHa-3D uses a coding agent to construct and refine editable Blender scenes from video~\citep{xu2026aha3d}, while LiteReality-Agent combines capture-view comparisons with geometric measurements and separate collision and support checks~\citep{huang2026literealityagent}.
CoDimRecon couples visual refinement with behavioral tests that guide revisions to deformable geometry and physical models~\citep{xie2026codimrecon}.
MUSE formulates scene authoring as incremental requirement satisfaction, using persistent memory to track progress and preserve content outside the edit target~\citep{xu2026muse}.

CoAgent3D integrates visual and geometric evidence into a relation-guided diagnosis--repair--review loop.
Mesh measurements, targeted cross-sections, and reference-view render comparisons guide the diagnosis of spatial inconsistencies.
Independent review evaluates the composed scene, updates scene relations and supporting evidence, and identifies unresolved or newly exposed issues for the next iteration.
% ScenePact Method: concise revision with an end-to-end pipeline and mechanism diagrams.
% Required packages: graphicx and the citation package already used by the paper.
% Figure files are relative to the main document directory.
% One sentence per source line; blank lines separate paragraphs.
\section{Method}\label{sec:method}

\begin{figure*}[t]
\centering
\includegraphics[width=\textwidth]{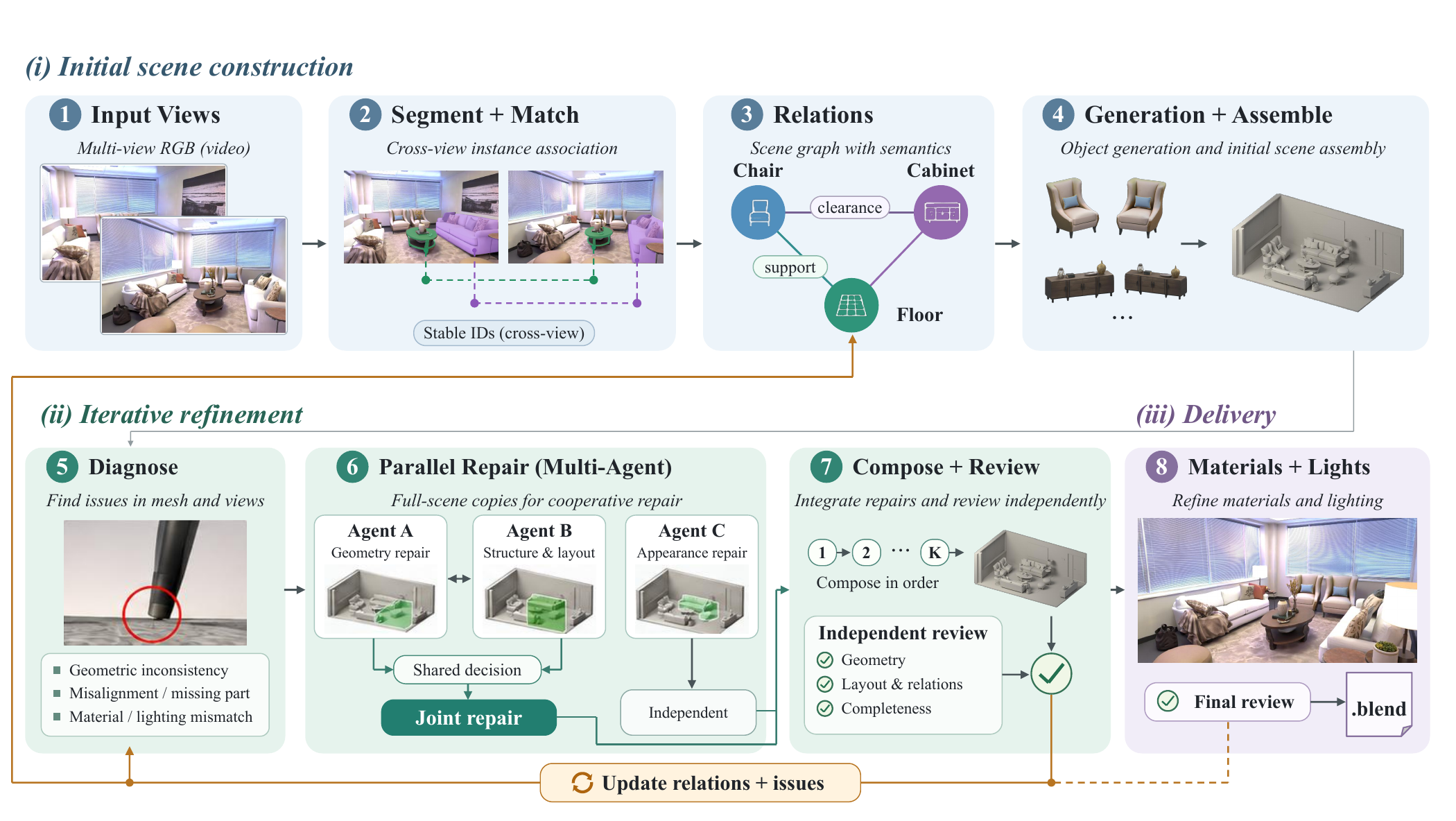}
\caption{\textbf{From reference images to an editable Blender scene.}
Cross-view instance association and relations inferred from observations guide object construction and placement.
The assembled scene enters a loop of diagnosis, cooperative repair, composition, and independent review.
Review updates relations and remaining issues; material and lighting refinement and final review produce the \texttt{.blend} output.}
\label{fig:framework}
\end{figure*}

CoAgent3D converts reference images into an editable Blender scene through \textbf{relation-guided cooperative repair}.
We associate observations with persistent object identities, establish scene relations, and assemble an initial scene.
Repeated diagnosis, parallel repair, and independent review then refine its geometry and appearance.
Figure~\ref{fig:framework} shows the complete process from input images to the final \texttt{.blend} scene.

\subsection{From observations to scene relations}\label{sec:state}

We collect reference images with any supplied cameras and depth, estimate missing quantities with Pi3X~\cite{wang2025pi3,pi3x2025software}, and establish a common world coordinate system.
Image resizing and cropping are propagated to masks, depth, and camera intrinsics.
An agent first examines the images to discover objects and architectural regions, then uses SAM~3~\cite{carion2025sam3} to obtain segmentation masks.
It associates these observations across views using appearance, co-visibility, tracking, spatial context, and reliable reprojection.
Each physical instance receives a stable identifier linking its image regions, visible parts, semantic orientation, and supporting observations.

These instances become nodes of an \textbf{object-relation graph}.
The agent constructs edges from the observed arrangement, recording the participating instances and parts, reference frame, source views, and expected orientation, support, contact, attachment, containment, or clearance.
For example, observations can require a chair foot to contact the floor before a corresponding mesh exists.
Each relation is marked as observed, inferred, or unresolved and retains its evidence.
After scene assembly, mesh measurements record how the reconstruction satisfies these requirements.
New evidence and corrected instance associations update the same graph throughout reconstruction.

\subsection{Initial construction and diagnosis}\label{sec:initialization}

We construct architectural surfaces in Blender from the scene observations and create separate object assets from their associated visual evidence.
For each object, the pipeline prepares a reference-conditioned base image, generates four directional views from that base, extracts the foreground, and invokes Generative Structure from Motion (GenSfM)~\cite{duan2026gensfm} to generate the asset.
Assets are scaled, oriented, and positioned in shared world coordinates using their observations and scene relations.
We bind each physical instance to its Blender mesh parts, yielding an editable initial scene with geometry, transforms, textures, and inspection cameras.

\begin{figure*}[t]
\centering
\includegraphics[width=\textwidth]{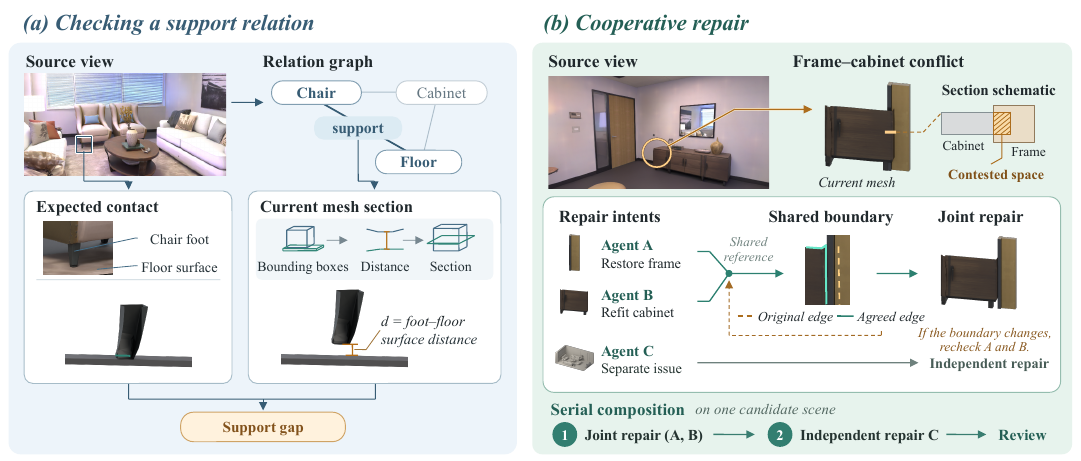}
\caption{\textbf{Relations connect diagnosis and cooperation.}
(a) Observed support defines a requirement; the current mesh and its cross-section reveal the gap to repair.
(b) Agents A and B inspect complete scene copies and agree on the boundary between a frame and a cabinet, producing a joint repair.
Agent C continues independent preparation.
Composed repairs undergo review, whose evidence updates subsequent decisions.}
\label{fig:relations-repair}
\end{figure*}

We compare this assembly with the observations and inspect orientation, support, contact, and clearance or interpenetration, in that order.
Bounding boxes select candidate object pairs; triangle intersections localize crossings, and mesh-surface distances reveal support and contact gaps.
These locations guide targeted mesh-plane cross-sections, complemented by renders from the reference cameras.
Comparable before-and-after sections expose local changes, while renewed intersection queries check the edited region.
Each discrepancy becomes a repair issue with affected instances, relation references, evidence, and checks to perform after editing.
Scanning all instances and source regions also reveals missing objects and relations.
Figure~\ref{fig:relations-repair}(a) illustrates how an observed relation exposes a defect in the current mesh.

\begin{table}[t]
    \centering
    \caption{Tools used by CoAgent3D for scene reconstruction and appearance refinement.}
    \label{tab:scene-tools}
    \small
    \setlength{\tabcolsep}{4pt}
    \renewcommand{\arraystretch}{1.08}
    \begin{tabularx}{\columnwidth}{@{}l>{\raggedright\arraybackslash}X@{}}
        \toprule
        Tool & Purpose \\
        \midrule
        \texttt{Pi3X}~\citep{pi3x2025software}
            & Estimate missing camera parameters and depth from input views. \\
        \texttt{SAM 3}~\citep{carion2025sam3}
            & Segment objects in input images and extract foreground masks for generated views. \\
        \texttt{GenSfM}~\citep{duan2026gensfm}
            & Build a textured object mesh from four generated views. \\
        \texttt{SuperMat}~\citep{hong2025supermat}
            & Estimate surface roughness and metallicity while preserving base color. \\
        \texttt{LUMEN}~\citep{chen2026lumen}
            & Predict HDR lighting at scene probes to guide lighting fits in Blender. \\
        \texttt{Blender}
            & Model walls and floors, place assets, execute repairs, and render the scene. \\
        \bottomrule
    \end{tabularx}
\end{table}

\subsection{Cooperative repair with full-scene context}\label{sec:cooperation}
\label{sec:preparation}

Each repair agent receives a private copy of the complete input scene and access to all observations, cameras, relations, and previous evidence.
Issue assignments specify responsibility, and agents can inspect and modify related objects wherever a repair requires them.
Each agent prepares a complete repair program together with its target issues, assets, and evidence.

During preparation, agents publish \textbf{repair intents}: proposed changes, referenced scene elements, and open questions.
The runtime routes these intents to agents with overlapping references and to explicitly named recipients.
Agents use the graph and shared evidence to identify decisions that their repairs need to resolve together.
For example, restoring a wooden frame may require space occupied by an adjacent cabinet.
The two agents examine the same contours and cross-sections, determine their common boundary, and agree on the corresponding geometry changes (Figure~\ref{fig:relations-repair}(b)).

These agreements form \textbf{shared decisions}, specifying the common interpretation, evidence, applicability, and participating repairs.
A repair records the exact decision version it adopts; changes to that decision trigger re-examination by the affected agents.
Agents continue preparing independent work while awaiting a particular decision or tool result.

When edits are coupled, the original agents develop a \textbf{joint repair}.
They either agree on an ordered group of complete programs or produce an integrated program replacing the coupled repairs.
The group is treated as one unit during composition, preserving the operations that must be evaluated together.
Agents can initiate consultation directly; cycles in recorded dependencies or decision waits also bring the involved agents together.
Cooperation follows the dependencies of concrete repairs while each participant retains complete scene context.

\subsection{Composition, review, and scene output}\label{sec:repair}

Once the selected repairs are ready, we verify their adopted decisions and order the complete programs according to the resolved dependencies.
The executor applies them sequentially to one candidate scene, preserving each joint repair's agreed internal order.
It then saves, reopens, and renders the composed scene.

A reviewer distinct from the contributing agents receives the source observations and candidate renders.
Local review checks the original issues and affected neighboring relations; whole-scene review also examines unchanged objects and source regions for remaining discrepancies.
The resulting evidence updates the relation graph, closes resolved issues, and records unresolved or newly discovered issues for the next round.
Related checks are renewed after changes to geometry, cameras, or lighting.
The loop continues until the scene passes review or the repair budget is exhausted.

\begin{algorithm}[t]
\caption{Cooperative scene repair and review loop}
\label{alg:cooperative-repair}
\small
\algrenewcommand{\algorithmicindent}{1em}

\noindent\textbf{Inputs:} Scene $S$, observations $O$, relation graph $G$, issue ledger $L$.\par
\noindent\textbf{Limits:} $B$ rounds left, $K_{\max}$ retries per issue per round.\par
\noindent\textbf{Notation:} $W$ lists issue IDs; $k_i$ counts retries left for issue $i$ this round.
$E(i,L,k)$ holds when $i \in \operatorname{dom}(k)$, $k_i>0$, and the latest whole-scene review in $L$ flags $i$ for rework.\par

\begin{algorithmic}[1]
    \State $(W,L) \gets \Call{Diagnose}{S,O,G,L}$

    \Repeat
        \If{$W \ne \varnothing$ \textbf{and} $B > 0$}
            \State $B \gets B - 1$
            \State $k \gets \{\, i \mapsto K_{\max} \mid i \in W \,\}$
            \State $S \gets \Call{RepairAndCompose}{S,O,G,W}$

            \Loop
                \State $(W,L) \gets \Call{LocalReview}{S,O,G,L}$
                \State $W \gets \{\, i \in W \mid k_i > 0 \,\}$

                \If{$W = \varnothing$}
                    \State $(W,L) \gets \Call{ReviewScene}{S,O,G,L}$
                    \If{$\exists\, i \in W \text{ such that } E(i,L,k)$}
                        \State $W \gets \{\, i \in W \mid E(i,L,k) \,\}$
                    \Else
                        \State \textbf{break}
                    \EndIf
                \EndIf

                \State $k_i \gets k_i - 1 \quad \forall\, i \in W$
                \State $S \gets \Call{RepairAndCompose}{S,O,G,W}$
            \EndLoop
        \Else
            \State $(W,L) \gets \Call{ReviewScene}{S,O,G,L}$
        \EndIf
    \Until{$W = \varnothing$ \textbf{or} $B = 0$}

    \State \Return $(S,G,L,B)$
\end{algorithmic}
\end{algorithm}

We then refine material roughness and metallicity while retaining object textures, and fit shared illumination to the reference images.
Final review checks the resulting appearance and returns remaining discrepancies to the same repair loop within the remaining budget.
The output is an editable \texttt{.blend} scene containing object meshes, transforms, textures, materials, lights, and cameras, accompanied by the updated relation graph and review record.

% Authored section: evaluation
\section{Experiments}\label{sec:evaluation}

\begin{table*}[!t]
    \centering
    \caption{Geometry, input-view rendering, and novel-view rendering on
    \textbf{Replica} and \textbf{ScanNet}. Bold indicates the best result
    for each metric within each dataset.}
    \label{tab:main-results}
    \small
    \setlength{\tabcolsep}{4pt}
    \renewcommand{\arraystretch}{1.1}
    \begin{tabular*}{\textwidth}{@{\extracolsep{\fill}}llccccccc@{}}
        \toprule
        & & \multicolumn{2}{c}{\textbf{Geometry}}
          & \multicolumn{4}{c}{\textbf{Input views}}
          & \multicolumn{1}{c}{\textbf{Novel views}} \\
        \cmidrule(lr){3-4}\cmidrule(lr){5-8}\cmidrule(lr){9-9}
        \textbf{Dataset} & \textbf{Method}
          & CD$\downarrow$ & F1$\uparrow$
          & PSNR$\uparrow$ & SSIM$\uparrow$ & LPIPS$\downarrow$
          & DINOv3$\uparrow$ & DINOv3$\uparrow$ \\
        \midrule
        \multirow{5}{*}{\textbf{Replica}}
          & ReplicateAnyScene & 1.025 & 72.09 & 13.86 & 0.6367 & 0.5732 & 0.5681 & 0.5767 \\
          & HoloScene         & 0.841 & 77.40 & 18.33 & 0.6803 & 0.4792 & 0.3631 & 0.4018 \\
          & GPT-6.2 Astra     & 0.525 & 86.13 & 15.84 & 0.6394 & 0.4691 & 0.7206 & 0.6412 \\
          & AHa-3D            & 0.395 & 90.49 & 16.88 & 0.6565 & 0.4241 & 0.7327 & 0.6603 \\
          & Ours              & \textbf{0.381} & \textbf{96.82} & \textbf{19.22}
                              & \textbf{0.6818} & \textbf{0.3837}
                              & \textbf{0.8231} & \textbf{0.7182} \\
        \midrule
        \multirow{5}{*}{\textbf{ScanNet}}
          & ReplicateAnyScene & 1.3203 & 55.05 &  9.53 & 0.4802 & 0.6412 & 0.5257 & 0.4342 \\
          & HoloScene         & 1.0492 & 56.01 & 13.49 & 0.5770 & 0.6051 & 0.5188 & 0.4583 \\
          & GPT-6.2 Astra     & 0.9141 & 62.69 & 13.41 & 0.5626 & 0.5528 & 0.7532 & 0.4838 \\
          & AHa-3D            & 0.7570 & 74.42 & 13.59 & 0.5799 & 0.5419 & 0.6991 & 0.4775 \\
          & Ours              & \textbf{0.2994} & \textbf{95.74} & \textbf{14.03}
                              & \textbf{0.5865} & \textbf{0.5053} & \textbf{0.8263} & \textbf{0.5969} \\
        \bottomrule
    \end{tabular*}
\end{table*}

\begin{table*}[!t]
    \centering
    \caption{Input-view and novel-view rendering on \textbf{Mip-NeRF~360}.
    Scores are averaged equally over the evaluated scenes.
    Bold indicates the best result in each column.}
    \label{tab:mipnerf360-results}
    \small
    \renewcommand{\arraystretch}{1.1}
    \begin{tabular*}{\textwidth}{@{\extracolsep{\fill}}lccccc@{}}
        \toprule
        & \multicolumn{4}{c}{\textbf{Input views}}
        & \textbf{Novel views} \\
        \cmidrule(lr){2-5}\cmidrule(lr){6-6}
        \textbf{Method}
        & PSNR (dB)$\uparrow$ & RGB SSIM$\uparrow$ & LPIPS-Alex$\downarrow$
        & DINOv3$\uparrow$ & DINOv3$\uparrow$ \\
        \midrule
        ReplicateAnyScene &  9.17 & 0.4974 & 0.7371 & 0.5134 & 0.4827 \\
        HoloScene         & 14.42 & 0.5754 & 0.6680 & 0.2655 & 0.2471 \\
        GPT-6.2 Astra     & 14.10 & 0.5351 & 0.5126 & 0.6637 & 0.6054 \\
        AHa-3D            & 14.38 & 0.5535 & 0.4734 & 0.6946 & 0.6206 \\
        Ours              & \textbf{16.20} & \textbf{0.5941} & \textbf{0.3867}
                          & \textbf{0.7578} & \textbf{0.6540} \\
        \bottomrule
    \end{tabular*}
\end{table*}

\begin{figure*}[!t]
    \centering
    {\footnotesize
    \begin{tabular}{@{}*{6}{c@{}}}
        \makebox[\dimexpr\linewidth/6\relax]{Reference} &
        \makebox[\dimexpr\linewidth/6\relax]{ReplicateAnyScene} &
        \makebox[\dimexpr\linewidth/6\relax]{HoloScene} &
        \makebox[\dimexpr\linewidth/6\relax]{GPT-6.2 Astra} &
        \makebox[\dimexpr\linewidth/6\relax]{AHa-3D} &
        \makebox[\dimexpr\linewidth/6\relax]{\textbf{Ours}}
    \end{tabular}\par}
    \vspace{2pt}
    \includegraphics[width=\linewidth]{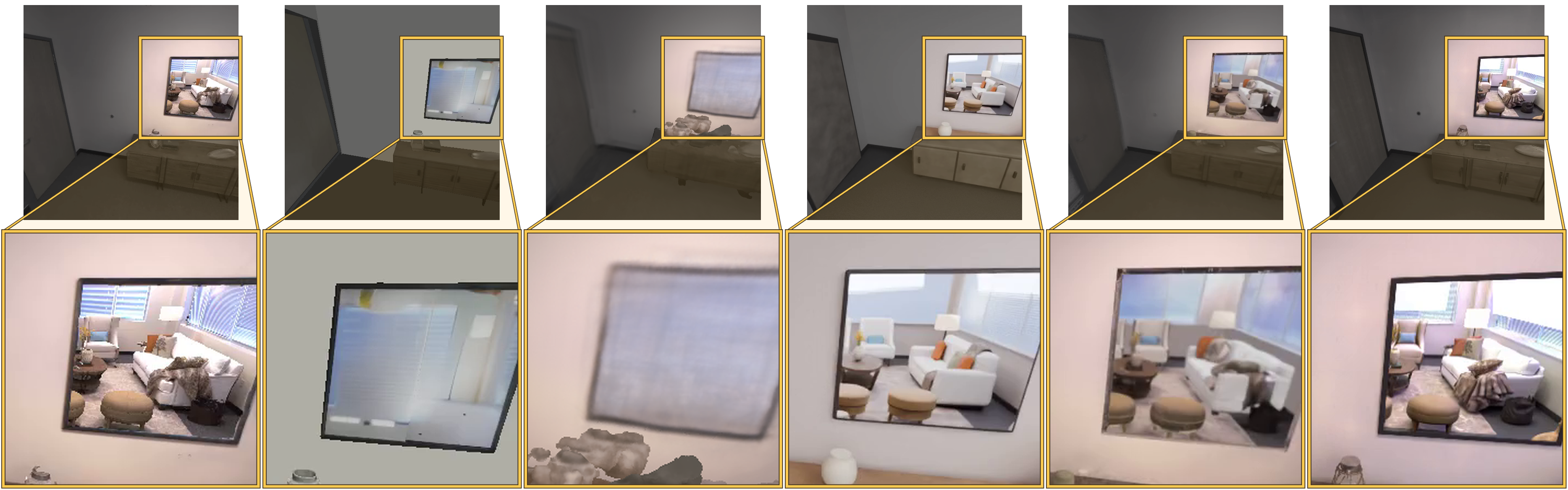}
    \caption{\textbf{Mirror detail in Replica \texttt{room\_0}.} The top row shows the reference frame and five reconstructions from the same camera view. The bottom row enlarges the mirror region outlined in yellow.}
    \label{fig:replica-mirror}
\end{figure*}

\begin{figure*}[!t]
    {\footnotesize
    \begin{tabular}{@{}*{3}{c@{}}}
        \makebox[\dimexpr\linewidth/3\relax]{Reference} &
        \makebox[\dimexpr\linewidth/3\relax]{ReplicateAnyScene} &
        \makebox[\dimexpr\linewidth/3\relax]{HoloScene}
    \end{tabular}\par}
    \vspace{2pt}
    \includegraphics[width=\linewidth]{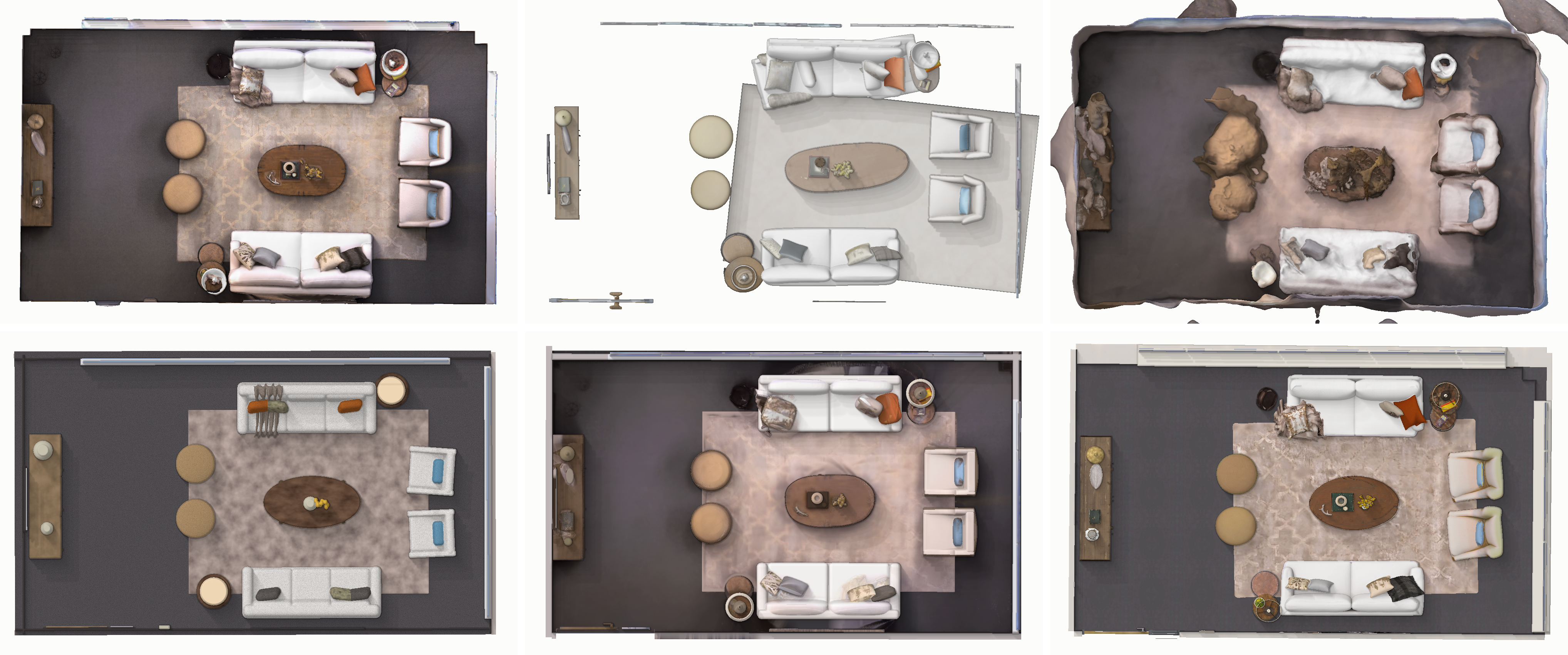}\par
    \vspace{2pt}
    {\footnotesize
    \begin{tabular}{@{}*{3}{c@{}}}
        \makebox[\dimexpr\linewidth/3\relax]{GPT-6.2 Astra} &
        \makebox[\dimexpr\linewidth/3\relax]{AHa-3D} &
        \makebox[\dimexpr\linewidth/3\relax]{\textbf{Ours}}
    \end{tabular}\par}
    \caption{\textbf{Top-down comparison on Replica \texttt{room\_0}.} The overhead view shows the arrangement of the rug and seating around the coffee table, the rug texture, and the shapes of the sofa cushions.}
    \label{fig:replica-top-view}
\end{figure*}

\begin{figure*}[!t]
    \centering
    \includegraphics[width=\textwidth]{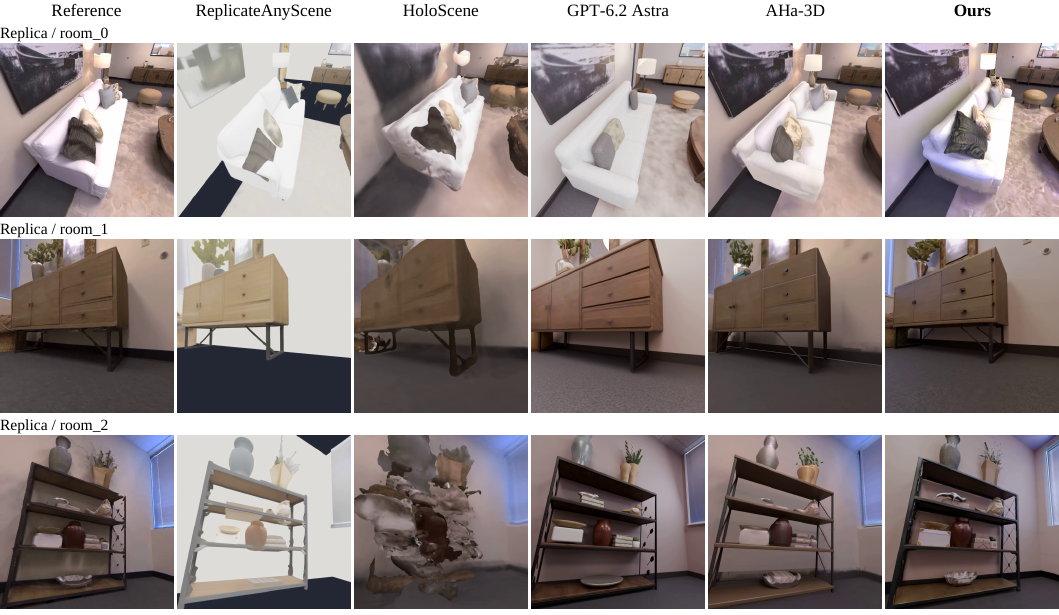}
    \caption{Input-view comparisons on Replica.}
    \label{fig:replica-input-views}
\end{figure*}

\begin{figure*}[!t]
    \centering
    \includegraphics[width=\textwidth]{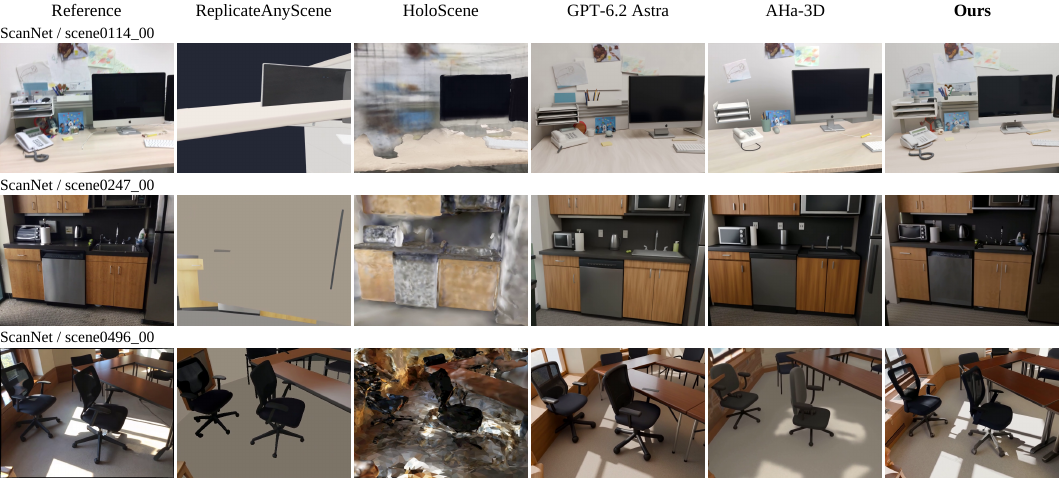}
    \caption{Input-view comparisons on ScanNet.}
    \label{fig:scannet-input-views}
\end{figure*}

\begin{figure*}[!t]
    \centering
    \includegraphics[width=\textwidth]{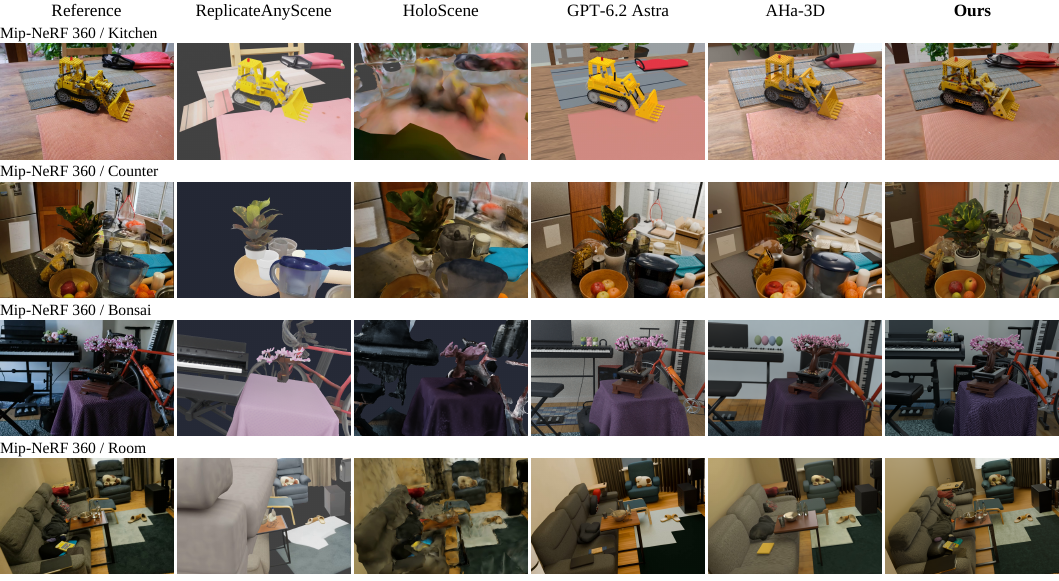}
    \caption{Input-view comparisons on Mip-NeRF~360.}
    \label{fig:mipnerf360-input-views}
\end{figure*}
\subsection{Setup}\label{sec:setup}

\noindent\hspace*{\parindent}\textbf{Datasets.}
We evaluate on ten indoor scenes across three datasets: three Replica~\citep{straub2019replica} scenes from the HoloScene release~\citep{xia2025holoscene}, three scenes randomly sampled from ScanNet~\citep{dai2017scannet}, and four scenes from Mip-NeRF~360~\citep{barron2022mipnerf360}.
We use one additional café scene from DL3DV~\citep{ling2023dl3dv10klargescalescenedataset} for the repair cooperation ablation.

\textbf{Baselines.}
We compare our approach with four baselines.
HoloScene~\citep{xia2025holoscene} reconstructs an initial mesh and infers a scene graph during geometric post-processing.
ReplicateAnyScene~\citep{dong2026replicateanyscene} is a zero-shot approach to compositional scene reconstruction.
For our comparison, we reimplement its pose-alignment module, which was omitted from the public release.
AHa-3D~\citep{xu2026aha3d} uses an agent to build and revise the scene in Blender.
To assess what our workflow adds beyond direct scene authoring, we also prompt GPT-6.2 Astra (hereafter Astra) to build the scene directly in Blender.
We set the reasoning effort to \texttt{xhigh} for all Astra calls.

\textbf{Inputs.}
For each scene, all methods start from the same RGB video.
Our method, AHa-3D, and the direct Astra baseline take the video as their sole input.
To keep the comparison consistent, we generate the auxiliary inputs required by HoloScene and ReplicateAnyScene from the same video using the same components that our agent uses internally.
Pi3X~\citep{pi3x2025software} estimates camera intrinsics, poses, and depth, while Astra identifies objects in keyframes and prompts SAM~3~\citep{carion2025sam3} to segment and track them across frames.
ReplicateAnyScene also receives per-pixel 3D point maps from Pi3X and category-level support and attachment relations inferred by Astra.

\textbf{Metrics.}
On Replica and ScanNet, we report Chamfer distance (CD) as a percentage of each scene's reference diagonal $D$ and compute F1 at a distance threshold of $0.01D$.
On all three datasets, we compare input-view renderings with their corresponding RGB frames using PSNR, SSIM~\citep{wang2004ssim}, LPIPS~\citep{zhang2018lpips}, and DINOv3~\citep{simeoni2025dinov3} feature similarity.
Following SceneAnything~\citep{xue2026sceneanything}, we score each novel-view rendering by its maximum DINOv3 similarity to the candidate frames from the input video.
For each rendering metric, we average scores over views within each scene, then average the scene scores with equal weight within each dataset.

\subsection{Quantitative Comparison}\label{sec:quantitative}

Our method ranks first on all seven metrics on both Replica and ScanNet (Table~\ref{tab:main-results}) and on all five metrics on Mip-NeRF~360 (Table~\ref{tab:mipnerf360-results}).
On Replica, our method reduces CD from 0.395 to 0.381 and raises F1 from 90.49\% to 96.82\% relative to AHa-3D.
Input-view SSIM is close to HoloScene's (0.6818 vs.\ 0.6803), while LPIPS is 9.5\% lower than AHa-3D's.
The F1 gain is larger on ScanNet, where our method exceeds AHa-3D by 21.32 percentage points (95.74\% vs.\ 74.42\%).
On the same dataset, novel-view DINOv3 similarity reaches 0.5969, compared with 0.4838 for GPT-6.2 Astra.
On Mip-NeRF~360, our method improves input-view PSNR by 1.78\,dB over HoloScene and reduces LPIPS by 18.3\% relative to AHa-3D.
It also improves novel-view DINOv3 similarity from 0.6206 to 0.6540 over AHa-3D.

\subsection{Qualitative Comparison}\label{sec:qualitative}

In the mirror crop (Fig.~\ref{fig:replica-mirror}), the throw over the sofa arm appears as a smooth patch in Astra, while AHa-3D captures its outline but leaves the folds indistinct.
Our rendering resolves the folds as the fabric falls over the armrest, as well as the slats of the blinds behind the sofa.

The overhead view (Fig.~\ref{fig:replica-top-view}) shows that our reconstruction preserves the alignment of the rug with the sofas and the gap between the two ottomans.
ReplicateAnyScene places the rug at an angle to the sofas, while the irregular ottoman contours in HoloScene obscure the gap.
Compared with Astra and AHa-3D, our reconstruction also more closely matches the fine tonal variation of the reference rug and the distinct shapes of the two overlapping cushions on the lower sofa.

\subsection{Ablation Studies}\label{sec:ablation}

\noindent\hspace*{\parindent}\textbf{Effect of iterative repair.}
Of 18 versions evaluated across three Replica rooms, we report each room's initial draft and five later checkpoints (Table~\ref{tab:replica-repair-loop}).
Loop~1 accounts for 1.94 of the total 2.57 percentage points gained in F1. From Loop~1 to Loop~5, PSNR improves by another 2.35\,dB and LPIPS falls from 0.4266 to 0.3837. Loop~5 achieves the best score on all seven metrics in Table~\ref{tab:replica-repair-loop}.

\begin{table}[!t]
    \centering
    \caption{Repair checkpoints on \textbf{Replica}, averaged equally over
    three rooms. Loop~0 is the initial draft and Loop~5 is the final
    reconstruction. }
    \label{tab:replica-repair-loop}
    \footnotesize
    \setlength{\tabcolsep}{1.5pt}
    \renewcommand{\arraystretch}{1.08}
    \begin{tabular*}{\columnwidth}{@{\extracolsep{\fill}}l*{7}{c}@{}}
        \toprule
        & \multicolumn{2}{c}{\textbf{Geometry}}
        & \multicolumn{4}{c}{\textbf{Input views}}
        & \textbf{Novel views} \\
        \cmidrule(lr){2-3}\cmidrule(lr){4-7}\cmidrule(l){8-8}
        \textbf{Loop}
        & CD$\downarrow$ & F1$\uparrow$
        & PSNR$\uparrow$ & SSIM$\uparrow$
        & LPIPS$\downarrow$ & DINOv3$\uparrow$ & DINOv3$\uparrow$ \\
        \midrule
        0 & 0.4356 & 94.25 & 15.15 & 0.6462 & 0.4526 & 0.7646 & 0.6494 \\
        1 & 0.4151 & 96.19 & 16.87 & 0.6598 & 0.4266 & 0.7940 & 0.6813 \\
        2 & 0.4157 & 96.28 & 17.90 & 0.6692 & 0.4069 & 0.8004 & 0.6823 \\
        3 & 0.4062 & 96.50 & 18.36 & 0.6713 & 0.3906 & 0.8216 & 0.7057 \\
        4 & 0.3987 & 96.71 & 18.90 & 0.6778 & 0.3855 & 0.8213 & 0.7137 \\
        5 & \textbf{0.3813} & \textbf{96.82} & \textbf{19.22}
          & \textbf{0.6818} & \textbf{0.3837} & \textbf{0.8231}
          & \textbf{0.7182} \\
        \bottomrule
    \end{tabular*}
\end{table}

\begin{figure*}[!t]
    \centering
    \includegraphics[width=\textwidth]{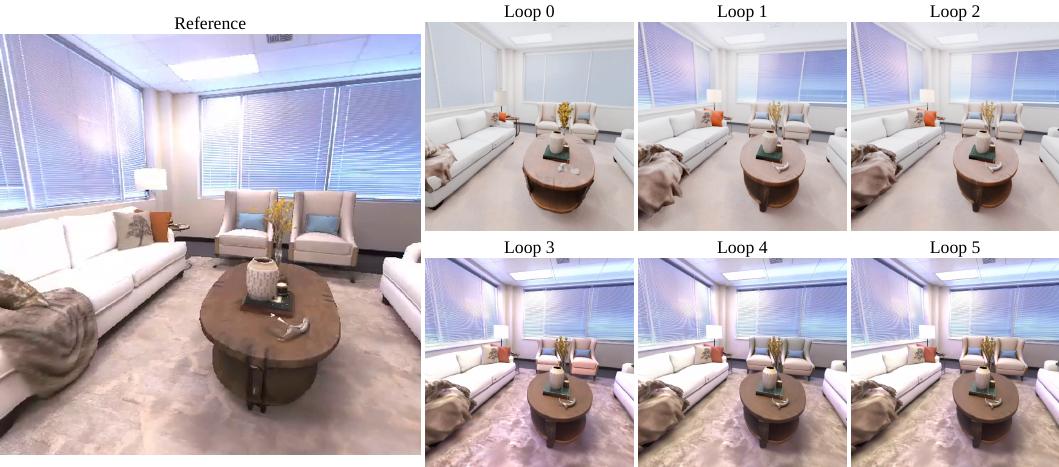}
   \caption{\textbf{Effect of iterative repair on Replica \texttt{room\_0}.}}
    \label{fig:ablation-replica-loops}
\end{figure*}

\begin{figure*}[!t]
    \centering
    \includegraphics[width=\textwidth]{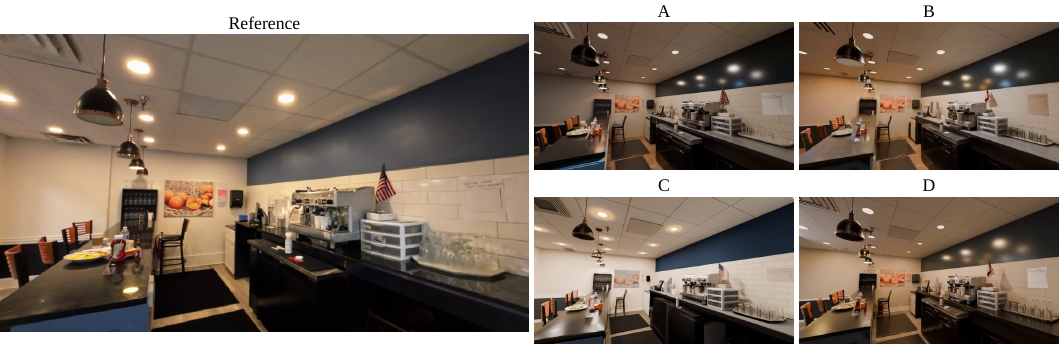}
    \caption{\textbf{Ablation comparison on the caf\'e scene.}}
    \label{fig:ablation-cafe-abcd}
\end{figure*}

\noindent\hspace*{\parindent}\textbf{Effect of the repair cooperation strategy.}
We compare four repair configurations to examine how scene context and
inter-agent coordination affect reconstruction quality and runtime:
\textbf{(A)} serial full-scene repair by a single agent;
\textbf{(B)} partitioned parallel repair, where each agent receives only
its assigned scene region;
\textbf{(C)} full-scene parallel repair without inter-agent negotiation;
and \textbf{(D)} CoAgent3D, where each agent prepares a repair plan on
a complete scene copy, broadcasts its dependencies using the relation
graph, and negotiates joint repairs with affected peers. We use 3 agents for all parallel configurations.

\begin{table}[!t]
    \centering
    \caption{Repair cooperation strategies on the café scene from
    \textbf{DL3DV}. Bold indicates the best result in each column.}
    \label{tab:coordination-ablation}
    \footnotesize
    \setlength{\tabcolsep}{1.5pt}
    \renewcommand{\arraystretch}{1.08}
    \begin{tabular*}{\columnwidth}{@{\extracolsep{\fill}}l*{6}{c}@{}}
        \toprule
        & \multicolumn{4}{c}{\textbf{Input views}}
        & \textbf{Novel views}
        & \multirow{2}{*}{\shortstack{\textbf{Runtime}$\downarrow$\\(h)}} \\
        \cmidrule(lr){2-5}\cmidrule(lr){6-6}
        \textbf{Setting}
        & PSNR$\uparrow$ & SSIM$\uparrow$ & LPIPS$\downarrow$
        & DINOv3$\uparrow$ & DINOv3$\uparrow$ & \\
        \midrule
        A & 13.84 & 0.4768 & 0.3464 & 0.7747 & 0.7222 & 15.1 \\
        B & 14.32 & 0.4874 & 0.4069 & 0.7098 & 0.7285 & \textbf{3.2} \\
        C & 13.65 & 0.4787 & 0.3679 & 0.6901 & 0.6925 & \textbf{3.2} \\
        \midrule
        D (Ours)
          & \textbf{16.39} & \textbf{0.5578} & \textbf{0.3145}
          & \textbf{0.8014} & \textbf{0.7431} & 3.3 \\
        \bottomrule
    \end{tabular*}
\end{table}

Table~\ref{tab:coordination-ablation} shows that CoAgent3D achieves
the best result on all five reconstruction-quality metrics.
Compared with serial repair (A), it increases PSNR by 2.55\,dB
and reduces LPIPS by 9.2\%, while reducing runtime from 15.1
to 3.3\,h, corresponding to a $4.58\times$ speedup.
Compared with the partitioned (B) and full-scene (C) parallel settings,
it improves PSNR by 2.07 and 2.74\,dB, respectively,
with runtime increasing by only 0.1\,h.
The advantage over (C) also extends to perceptual quality and
novel-view similarity: LPIPS decreases by 14.5\%, and novel-view
DINOv3 similarity increases from 0.6925 to 0.7431.
These gains demonstrate the benefit of our cooperative repair design
beyond full-scene access alone.
By making repair dependencies explicit and negotiating joint repairs,
CoAgent3D combines the efficiency of parallel preparation with
higher reconstruction quality than both serial and uncoordinated
parallel repair.

\section{Discussion and Limitations}\label{sec:limitations}

CoAgent3D uses visual observations and spatial relations to guide reconstruction and coordinate repair.
Reference images constrain visible geometry and appearance, while occluded structures are completed using generative priors and surrounding scene context.
Additional viewpoints and explicit uncertainty estimates offer a direction for resolving ambiguous regions and strengthening the evidence behind shared repair decisions.

Full-scene access and direct negotiation introduce computation and communication costs as scenes become larger and more interconnected.
Repair preparation proceeds in parallel, while complete repair programs are composed sequentially according to their resolved dependencies.
Reusing validated geometric evidence and reducing repeated inspection are practical directions for scaling cooperation to larger scenes.

The current diagnosis and review process focuses on visual fidelity and static spatial consistency.
Integrating physical simulation would extend the relation-guided feedback loop to physical stability and dynamic interaction, including articulation and deformation.
This extension would connect reconstruction of geometry and appearance with verification of scene behavior under interaction.

\section{Conclusion}\label{sec:conclusion}

We presented CoAgent3D, a relation-guided self-improvement framework for editable 3D scene reconstruction.
Every repair agent retains access to the complete scene and prepares edits in parallel, with cooperation organized around dependencies between proposed repairs.
Repair intents and shared decisions enable affected agents to negotiate coupled changes and develop joint repairs.
An evolving relation graph connects visual observations and geometric measurements to repair planning.
Independent review of the composed scene updates these relations and guides subsequent rounds of diagnosis and repair.
Across real-world scenes, CoAgent3D produces photorealistic reconstructions with complete, editable object geometry, textures, materials, and scene lighting.

\FloatBarrier
\clearpage
\bibliographystyle{unsrtnat}
\begingroup
\bibliography{references_verified}
\endgroup
\end{document}